\def\year{2022}\relax
\documentclass[letterpaper]{article} % DO NOT CHANGE THIS
\usepackage{aaai22}  % DO NOT CHANGE THIS
\usepackage{times}  % DO NOT CHANGE THIS
\usepackage{helvet}  % DO NOT CHANGE THIS
\usepackage{courier}  % DO NOT CHANGE THIS
\usepackage[hyphens]{url}  % DO NOT CHANGE THIS
\usepackage{graphicx} % DO NOT CHANGE THIS
\usepackage{natbib}  % DO NOT CHANGE THIS AND DO NOT ADD ANY OPTIONS TO IT
\usepackage{caption} % DO NOT CHANGE THIS AND DO NOT ADD ANY OPTIONS TO IT

\DeclareCaptionStyle{ruled}{labelfont=normalfont,labelsep=colon,strut=off} % DO NOT CHANGE THIS
\usepackage{algorithm}
\usepackage{algorithmic}
\usepackage{multirow}
\usepackage{array}
\usepackage{bm}
\usepackage{epstopdf}
\usepackage{subfigure}
\usepackage{amsmath,amssymb,amsfonts}

\usepackage{newfloat}
\usepackage{listings}
\floatstyle{ruled}
\newfloat{listing}{tb}{lst}{}
\floatname{listing}{Listing}
\title{DBT-DMAE:\\An Effective Multivariate Time Series Pre-Train Model under Missing Data}
\author{
    Kai Zhang, Qinmin Yang, Chao Li
}
\affiliations{
    \textsuperscript{\rm 1}Zhejiang University Control Engineering and Science Department\\
}

\usepackage{bibentry}
\begin{document}

    \maketitle

    \begin{abstract}
    Multivariate time series(MTS) is a universal data type related to many practical applications. However, MTS suffers from missing data problems, which leads to degradation or even collapse of the downstream tasks, such as prediction and classification. The concurrent missing data handling procedures could inevitably arouse the biased estimation and redundancy-training problem when encountering multiple downstream tasks. This paper presents a universally applicable MTS pre-train model, DBT-DMAE, to conquer the abovementioned obstacle. First, a missing representation module is designed by introducing dynamic positional embedding and random masking processing to characterize the missing symptom. Second, we proposed an auto-encoder structure to obtain the generalized MTS encoded representation utilizing an ameliorated TCN structure called dynamic-bidirectional-TCN as the basic unit, which integrates the dynamic kernel and time-fliping trick to draw temporal features effectively. Finally, the overall feed-in and loss strategy is established to ensure the adequate training of the whole model. Comparative experiment results manifest that the DBT-DMAE outperforms the other state-of-the-art methods in six real-world datasets and two different downstream tasks. Moreover, ablation and interpretability experiments are delivered to verify the validity of DBT-DMAE's substructures. 
    \end{abstract}

    \section{Introduction}
    \label{sec:introduction}
    	\begin{figure*}[htb]
    		\centering{\includegraphics[width=0.7\textwidth]{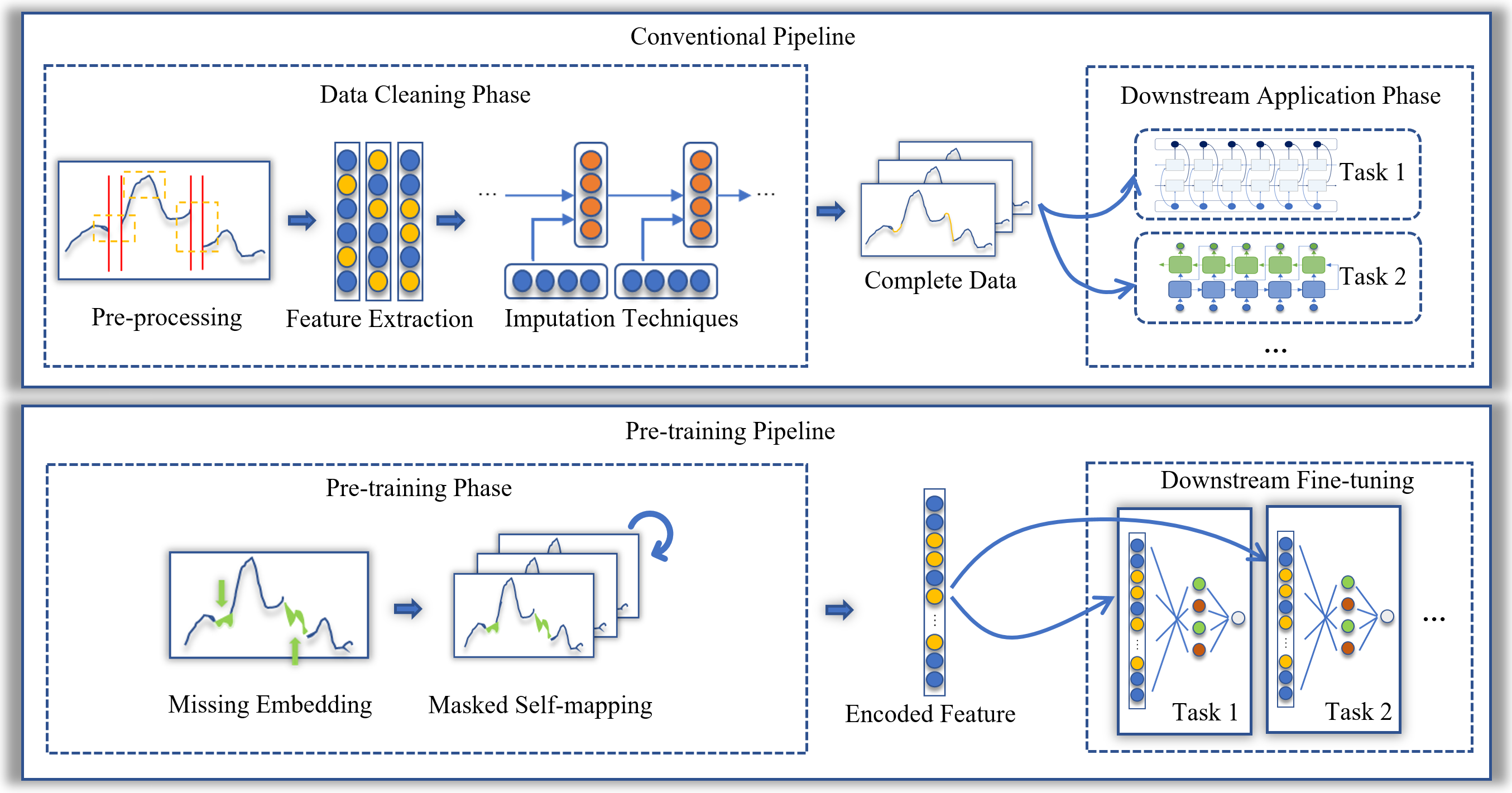}}
    		\caption{In conventional procedures, the missing data imputation is treated as the data pre-processing, so multiple downstream task models have to be trained individually. In our proposed pre-train designs, the fully-encoded generalized feature can be easily fine-tuned and oriented to diverse downstream tasks after introducing the missing embedding and masked self-mapping pre-training.}
    		\label{f_Head}
    	\end{figure*}
    
    	Multivariate time series(\textbf{MTS}) data have extensive practical applications in industry, health care, geoscience, biology, etc. However, MTS universally carries missing observations in diverse data contexts. It has been noted that these missing entries provide informative features of the original data sources\cite{rubin1976inference}. As a result, missing data's discriminative characteristics must be considered when  dealing with MTS-related tasks.
    	
    	In literature, the MTS-related tasks encountering missing data are often solved by the procedures shown in the upper part of Figure \ref{f_Head}. Under the two-phase pipeline, missing imputation is treated as the data-cleaning stage, and then independent models oriented towards multiple downstream tasks are conducted individually in the application stage. Many valuable works have contributed to missing data imputation in MTS under this framework. The methodologies cover many fields, including tensor decomposition, RNN-based prediction, auto-encoders, and generative-adversarial theory, etc. However, there are several drawbacks for present actual applications. The traditional therapies produce a complete dataset as an intermediate product which inevitably brings imputation bias to missing entries. Besides, the feature extracting procedures are repeatedly developed in both phases, and multiple downstream models have to be trained independently, which always results in  significant redundancy.
    	
    	Moreover, missing data in MTS has its unique missing patterns impairing the concurrent treatments mentioned above. Apart from the common patterns listed in Rubin's work\cite{rubin1976inference}, MTS missing data also confronts some reluctant patterns, often referring as "line missing" and "block missing,"\cite{cini2021filling} contributing to the disconnection of measurement within a global sensory integration during network no-response or downtime. These patterns deprive the MTS's time-continuous property, and the missing entries are jointly missing simultaneously. As a result, many conventional imputation methods mentioned above become invalid even under a relatively small missing ratio. 
    	
    	Based on these considerations and inspired by the pre-train idea in NLP and CV field, we novelly proposed a generalized MTS pre-train model called DBT-DMAE. Our model adopts a typical pre-train procedure depicted in the lower part of Figure \ref{f_Head}. In order to avoid the training redundancy problem caused by obtaining the complete dataset in the conventional pipeline, DBT-DMAE utilizes an auto-regressive architecture under a masked learning mechanism and directly learns from the unlabeled incomplete MTS data to get generic encoded MTS representation. As for handling the tricky missing pattern appearing in MTS, we propose a dynamic missing positional embedding(\textbf{DPE}) technique giving all missing entries with effective representations without bringing in extra imputation bias.
    	
    	For detailed implementation, we propose a TCN-based\cite{TCN} unit, called Dynamic Bidirectional TCN(\textbf{DBT}), as the basic encoder unit to capture temporal correlations from a bidirectional data context in MTS. The entire DMAE model is designed to extract multi-time scale features and perform effective deep fusion to obtain universally applicable encoded features. For the DMAE training progress, the specialized data-feed-in and loss strategy guarantees adequate training of all substructures, and a warm-up training trick is applied to accelerate and stabilize the convergence progress. After the pre-train phase, downstream-task-oriented fine-tuning can be quickly delivered by substituting the decoder of the well-trained DMAE. In this paper, we choose the multi-step prediction and MTS classification task as the downstream task examples, and the fine-tuning could rapidly converge in less than several epochs. In general, our main contributions can be concluded as follows:
    	
    	\begin{enumerate}
    		\item A novel MTS pre-train framework under missing data is proposed. Under this framework, the mentioned biased-imputation problem is avoided by applying dynamic missing mask tokens derived from extensive unlabeled MTS data. At the same time, the straightforward downstream task fine-tuning procedure directly solves the redundant training problem in the conventional pipeline. 
    		\item The proposed DBT-DMAE holds distinguished adaptability to dynamic time-varying MTS input. The dynamic intrinsic nature of DPE, DK, and ASF mechanisms enables the model to draw a profoundly underlying temporal correlation in multiple scales and both time orientations and gives a preferable generalization performance.
    		\item The pre-train effectiveness is evaluated by two downstream tasks under six real-world open datasets whose fields range from industry, climates, wearable devices, and speech recognition.
    	\end{enumerate}
    	
    	The rest of this article is organized as follows: Section \ref{sec:Notation and Problem Statement} describes the notation used in this article and the problem formulation of DBT-DMAE. Section \ref{DBT} introduces the DBT-DMAE in detail. Section \ref{Experiment} includes all the experiments, including comparative studies, ablation experiments, and model interpretation experiments. Finally, a concise conclusion is made at the end of the paper.
    \section{Related Works}
    \label{sec:Related_works}
    \subsection{Missing Data Imputation}
         Under deep learning background, there are mainly three types of methodology of missing data imputation: prediction-based, auto-encoder-based, and GAN-based methods. For prediction-based methods, some works\cite{Deepmdp,GRNNSGTM} transform the imputation into an MTS prediction problem while using RNN-based models to predict the missing value. Some other works\cite{che2018recurrent,BRITS,tang2020joint} integrate the missing prediction as an intermediate step in time series prediction. In terms of auto-encoder-based methods, some other works\cite{miranda2011reconstructing,6097083,9768200} take the missing parts as random noises and recover the missing value with the output of the delicately-designed auto-encoder. Moreover, with the recent advancement of generative adversarial theory, many works\cite{MAGAN,E2GAN,GAIN} follow the basic generative adversarial idea with the utilization of deep learning neural networks to train the specifically structured generators and discriminators and generate the value of the missing parts.
    \subsection{Pre-train Models}
        In 2016, the Google Brain research team proposed the seq2seq pre-train model\cite{Seq2Seq}. Next up, in 2018, BERT\cite{BERT} was carried out by the Google AI Language research group and GPT\cite{GPT} by the OpenAI research team in the same year. Also, GPT-v2\cite{radford_language_2019} and GPT-v3\cite{brown_language_2020} were proposed in succession in the next few years. Furthermore, in 2021, the Facebook AI Research team led by Kaiming proposed the Masked Autoencoder\cite{MAE} pre-train model in CV.

	\section{Notation and Problem Statement}
	\label{sec:Notation and Problem Statement}
	Given MTS, $i.e.$, $X=(x^1, x^2,...,x^n)=(x_1, x_2, ..., x_T)\in R^{n\times T}$, where $T$ is the length of the sequence, we use $x^k=(x_1^k, x_2^k, ..., x_n^k)\in R^{T}$ to denote the kth attribute of length $T$ and employ $x_t=(x_t^1, x_t^2, ..., x_t^n)\in R^n$ to denote the attribute vector at time-entry $t$. Meanwhile, due to the missing phenomenon overwhelming in MTS data, we also introduce the binary missing mask matrix $M$ as the same shape as the $X$, in which $M_t^k=1$ indicates the $k_{th}$ attribute at time-entry $t$ is a valid value, while $M_t^k=0$ indicates the value is missing.
	
	Typically, in our work, DBT-DMAE implements the n-attribute sequential input self-mapping. Given the input and mask matrix $X, M\in R^{n\times T}$, the DBT model intends to project the input $X$ into a hidden representation $H\in R^m$ through an encoder $Enc$, where $m$ represents the number of hidden states, and reconstruct the complete $X$ by a decoder $Dec$ :

	\begin{equation}
			H = Enc(X,M)
	\end{equation}
	\begin{equation}
	    \hat{X}=Dec(H) 
	\end{equation}
    where $\hat{X}$ denotes the reconstructed result. 
    
	Apart from the DBT-DMAE problem statement, different downstream tasks have independent problem expressions. In this paper, we only refer to two types of downstream tasks:  multi-step prediction and classification. The followings are the problem statement for each task.
	
	\subsubsection{Multi-step Prediction Problem Statement}
	Prediction is a common task among all the MTS applications. We apply 1-step, 3-step, and 5-step prediction experiments in this work. For 1-step prediction, the exact next target value is predicted; for 3-step and 5-step prediction, the target values of the following three or five-time points are estimated. To be more specific, we expect to establish the functional relationship between target value $X^{target}$ and previously observed data $X$ as follows:
	\begin{equation}
		\hat{X}_{T+1:T+s}^{target}=f_p(X, M)=Dec_{p}(Enc(X,M))
	\end{equation}
	where $s$ denotes the prediction steps.

	\subsubsection{MTS Classification Problem Statement}
	MTS Classification is a substantial task with applications in many fields, including but not limited to wearable devices, monitoring systems, and speech recognition. Here, our objective is to find a non-linear multivariate probability distribution $f_c(c_i|X,M)$, which takes MTS data $X$ and mask matrix $M$ as input and outputs the probability that this series belongs to each class $c_i$.
	
    \begin{gather}
        p(c_i) = f_c(c_i|X, M) = Dec_{c}(c_i|Enc(X,M))\\
        \hat{c} = \arg\max_{c_i}p(c_i)
    \end{gather}

	\section{Model Architecture}
	\label{DBT}
	Dynamic Bidirectional Temporal Convolution Network-based Denoising Mask Auto-encoder, abbreviated to DBT-DMAE, is fully discussed in this section. The fundamental DBT unit and DBT block is first  introduced. Then, the specially designed DPE mechanism is enclosed to conduct the missing data representation.  At last, the overall DMAE structure is beheld with the data-feed-in and loss strategy, while the warm-up training trick is also explicitly included. 
	
	\subsection{Dynamic Bidirectional TCN Unit and Block}
	
	\begin{figure}[htb]
		\centering
		\subfigtopskip=-5pt
		% \subfigbottomskip=2pt
		% \subfigcapskip=-5pt

		\includegraphics[width=\columnwidth]{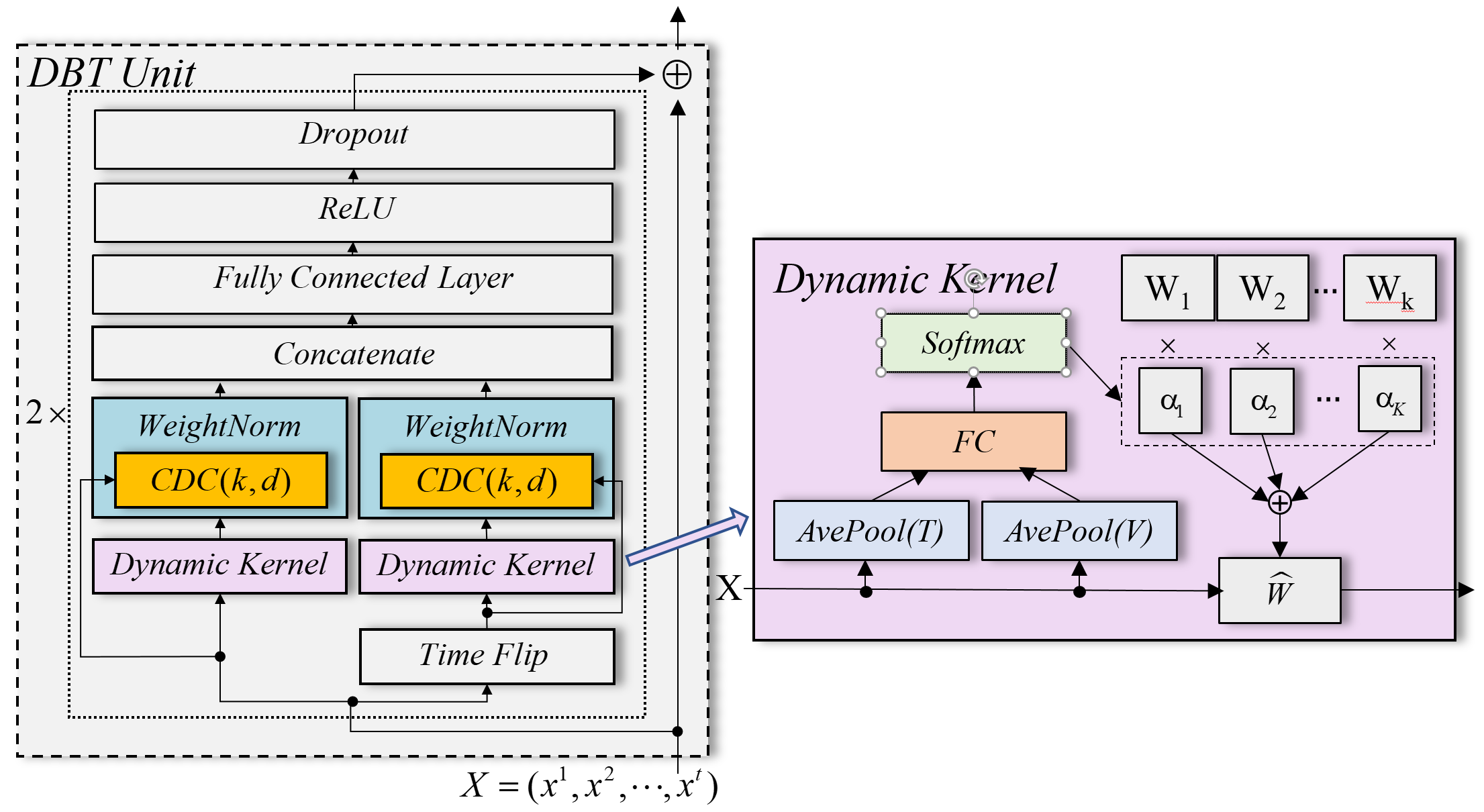}
		
		\caption{DBT: two-directional features are fused between each two adjacent CDC layers, and the dynamic kernel is adaptive to mutative input.}
		\label{f_TCN_structure}
	\end{figure}
	
	\begin{figure*}[htb]
		\centering
		\includegraphics[width=0.88\textwidth]{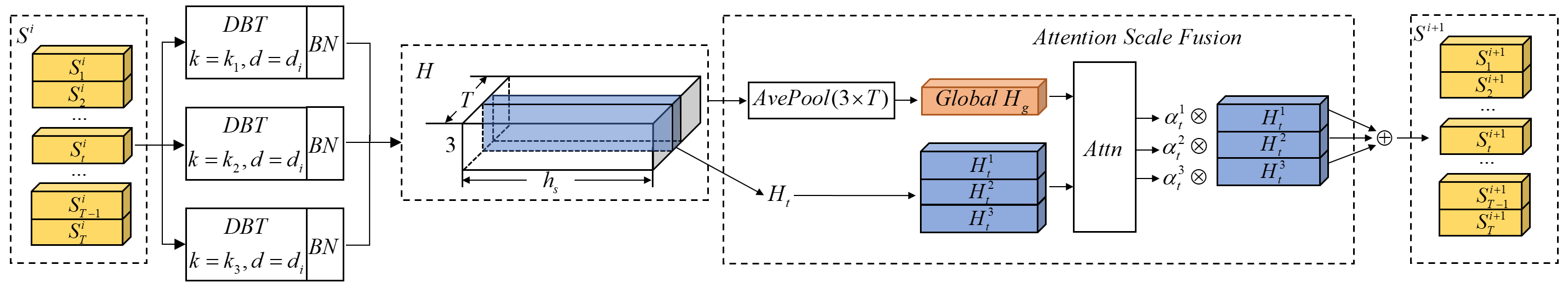}
		\caption{DBT block: an independent computing module that consists of three parallel DBT Units and an Attention Scale Fusion(ASF) Layer to fuse multi-scale features.}
		\label{f_DBT_Block}
	\end{figure*}
	
	\label{subsec:Dynamic-TCN}
	The dynamic bidirectional TCN unit(DBT) is an ameliorated TCN unit used as the basic unit of the DBT block, which will be included in this section later. The original TCN is a residual architecture with two sequentially stacked Casual Dilated Convolutional(CDC) layers and a ReLU non-linear activation. Based on this structure, we make two main modifications to the original TCN in general, whose interior structure is shown in Figure \ref{f_TCN_structure}.

    First, since our pre-train model is made to learn the underlying temporal relations within MTS, it is reasonable to integrate sequential information from both time directions. As a result, we apply the "time flipping" trick to the input MTS. As shown in Figure \ref{f_TCN_structure}, we adopt two independent CDC networks corresponding to time-forward and time-backward convolution. Afterward, a fully-connected layer merges information from opposite directions and maintains the appropriate shape of the next CDC layer required.
    
    Second, MTS's multi-variable characteristic leads to a wide range of input fields containing mutative conditions, and therefore a fixed combination of reception kernels could be insufficient. Based on this consideration, we propose the dynamic kernel(DK) to replace the convolution kernel in CDC layers. The detail is depicted on the right side of Figure \ref{f_TCN_structure}. Inspired by the idea of the work of Chen, Y.\cite{DynamicConv}, we use the input attention mechanism to fuse multiple learnable kernel groups into one. Specifically, we apply two average pooling upon the input $X$ squeezing the input length $T$ and variable dimension to obtain global vector entries. Then a linear projector and a SoftMax layer are employed on the global entries to get $K$ attention weights $\alpha_{k}(x)$. At last, the output kernel parameter is aggregated from $K$ groups of the kernel with the $K$ weights. The calculation process can be expressed as Equal.(\ref{eq_kernel_first})-(\ref{eq_kernel_last}), where $\mathbf{1}$ indicates the unit vector with all entries equal to 1; $\hat{W_c}$ is the output kernel arguments; $W_i$ is the weights of the $i$th candidate kernel group; $\lambda>1$ is the penalty factor to alleviate the one-hot phenomenon in softmax. Dimensions of all the parameters are coherent with the input $X$, which is easy to obtain.

	\begin{gather}
		\label{eq_kernel_first}
		\boldsymbol{a_d}=(a_1,a_2,...,a_K) = W_d[\frac{1}{T}X\mathbf{1}, \frac{1}{n}X^T\mathbf{1}]+b\\
		\label{eq_softmax_dk}
		\boldsymbol{\alpha_d}=(\alpha_1,\alpha_2,...,\alpha_K),\alpha_i=\frac{exp({a_i/\lambda})}{\sum_{i=0}^{K}exp({a_i/\lambda})}\\
		\hat{W_c}=\sum_{i=0}^{K}\alpha _i \cdot W_i 
		\label{eq_kernel_last}
	\end{gather}
	
	DBT block functions as an independent computing module in the overall model. One DBT block consists of three parallel DBT units with BatchNorm and an Attention Scale Fusion(ASF) layer incorporating all temporal features. The exact structure is depicted in Figure \ref{f_DBT_Block}. The three parallel DBT units have different kernel sizes to capture features in multiple sequence scales. After three feedforward paths, an ASF layer is attached to integrate features under different scales. Firstly, we draw an average pooling on the concatenated parallel output, only maintaining the hidden-feature $h_s$ dimension and squeezing the other dimensions to 1. Then, the average vector is used as a reference entry toward each $H_t$ of shape ($h_s$ $\times$ 3). The following attention function assigns each $H_t^i$ a weight $\alpha_t^i$, and the final output is computed as the inner product between the weight vector and the $H_t$. The Attention function we use here is calculated as follows:
	
	\begin{gather}
		e_t^i=v^{T}tanh(h_gW_g+H_t^iW_H)\\
		\label{eq_softmax_asf}
		\alpha_t^i=\frac{exp(e_t^i/\gamma)}{\sum_{i=1}^3exp(e_t^i/\gamma)} \\
		S_t=\sum_{i=1}^{3}\alpha_t^iH_t^i 
	\end{gather}
	where $W_g \in R^{h_s\times h_a}$ is the global feature projection operator($h_a$ is the attention hidden states); $\gamma > 1$ is also the penalty factor; $W_H\in R^{h_s\times h_a}$ is the local feature projection operator; $v\in R^{h_a\times 1}$ is the attention vector; the $S_t$ represents the final output of DBT Block at time t.

	\subsection{Dynamic Positional Embedding}
		\label{subsec:input representation}
	Without the fully imputed MTS as the input, the original MTS must be well represented in the proposed pipeline. Learning from the masked language modeling\cite{liu2021pre} coming up with numerous famous NLP pre-train models \cite{BERT,liu_roberta_2019}, we apply a random mask on the original MTS input and treat the masked entries equivalent to missing parts that are required to be restored. Unlike the mask token embedding method using the hard-coded or fully learnable embedding in NLP pre-train models, we consider that the missing token representation should combine time-varying and variable-varying characteristics, which means that the missing data appears at different time or variables would have discrepant token representations. As a result, we adopt a specialized end-to-end missing positional embedding technique, called dynamic positional embedding(DPE), to generate the time-variable-varying missing tokens. 
	
	Specifically, DPE is implemented by a random masking(RM) procedure and a DBT block. Apart from the initially missing entries, observations of a masking ratio are artificially masked, and both the masked and missing entries are embedded by the bidirectional scanning of a DBT block upon the masked data. Furthermore, the DBT unit is an end-to-end convolution architecture so that it has the representation ability to generate multiple positional embeddings at only one glance and could theoretically bear and process any missing pattern, including the "line missing" and "block missing" mentioned above. This whole DPE process can be formulated as Equal.(\ref{first_eq_dpe})-(\ref{last_eq_dpe})
	
	\begin{gather}
	    \label{first_eq_dpe}
	    X_p = X + p(\theta)\\
	    M_m = downsample(M) \\
	    \widetilde{X} = DBT(X_p \circ M_m)\\
	    X'=M_m \circ X_p + (\mathbf{1}-M_m) \circ \widetilde{X}
	    \label{last_eq_dpe}
	\end{gather}
	where $p$ is a random noise strengthening the robustness of the model with $\theta$ as its parameter; $downsample$ denotes the artificial masking function, which randomly sets zeros within the original missing mask matrix;$\circ$ is the element-wise multiplication.
	
	\subsection{Auto-Encoder Architecture}
	\label{subsec:Autoencoder Architecture}	
	
	\begin{figure}[h]
		\centering
		\includegraphics[width=1\linewidth]{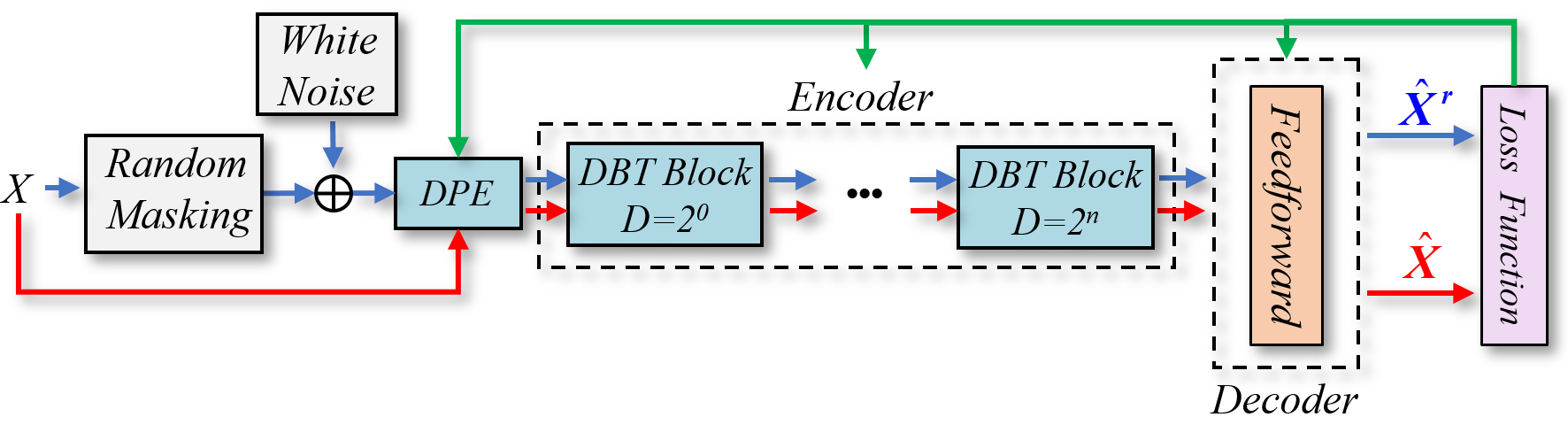}
		\caption{The overall architecture of DBT DMAE.}
		\label{f_Dynamic-BiTCN-MAE}
	\end{figure}
	The entire DBT-DMAE architecture is designed to establish an auto-regressive projection with valid missing entry representation under the missing data problem in the MTS. A DBT-based encoder and feedforward decoder constitute the holistic pre-train model. 
	
	The encoder consists of sequential DBT blocks. Within each block, all DBT units share the same dilation size, while the stacked blocks have gradually enlarged dilate sizes to integrate local and global temporal features in MTS. In practice, we use three consecutive DBT blocks stacked on top of each other, and their dilation size grows exponentially with base two. In terms of the decoder, a more complicatedly-structured decoder may contribute to a better performance of the auto-regressive reconstruction. However, in DBT-DMAE, we desire to obtain a generalized feature representation of the input MTS. With that being the case, a simple straight decoder is ideal for its low expressiveness, and thus a simple fully-connected feedforward network is introduced as the decoder in our work. The overall architecture of DBT is depicted in Figure \ref{f_Dynamic-BiTCN-MAE}.
	
	\subsection{Training Strategy}
	 It is worth mentioning that both original MTS and the one after RM are fed into the model. The red and blue lines in Figure \ref{f_Dynamic-BiTCN-MAE} denote the two forward paths, and the reconstructed results are denoted as $\hat{X}$ and $\hat{X}^r$, respectively. The objective function is as Equal.(\ref{eq_objective}), where $M$ is the missing mask matrix while $M^r$ is the mask matrix after RM; $m_r$ denotes the random masking ratio. The first part of the function focuses on the manually masked part, while the second part pays attention to reconstructing all observed data. This combination is critical because combining the two parts enforces DBT to learn  both the original data distribution and the missing representation at the same missing entry. Additionally, the two losses are weighted-combined corresponding to each missing ratio, which aims to balance the probabilities between the model seeing the artificially masked missing embedding and the corresponding original observed data. Furthermore, it has been proved by our ablation experiments that the pre-train performance would remarkably margin if the first part vanish.
	
	\begin{equation}
		\begin{aligned}
			\mathrm{Loss} =& \frac{1}{m_r+1} \frac{\left \|  (\hat{X}^r -X)\circ (M-M^r)  \right \|_2}{\left \|  M-M^r \right \|_2} \\&+\frac{m_r}{m_r+1} \frac{\left \|  (\hat{X} -X)\circ  M \right \|_2}{\left \|  M \right \|_2}
		\end{aligned}
		\label{eq_objective}
	\end{equation}
	
	For more detail of the training process, noticing the heavy utilization of softmax operation in the DBT unit and the sequentially stacked DMAE structure, direct training of the entire model could lead to oscillation, and the converge speed is slow. Therefore, we proposed a so-called "warm-up" training trick to give the network a better initialization firsthand. In several beginning epochs, we artificially replace all softmax with uniformed weights to neutralize the underfitting problem brought by the nearly one-hot output of softmax. In this process, all the model structures, especially all candidate kernels in DK, are sufficiently initialized to a relative ideal position. After several epochs' warm-up, the softmax layers then come into play. Also, the penalty factors used in Equal.(\ref{eq_softmax_dk}) and Equal.(\ref{eq_softmax_asf}) are set to a relatively large positive number to avert the sudden saltation within the model once the softmax resumes work.
	
	\subsection{Downstream Fine-tuning}
	As for the fine-tuning detail, in this paper, we include multi-step prediction and classification task fine-tuning. DBT-DMAE's original decoder is substituted for prediction and classification tasks with a new feedforward network. The last layer's output dimension is set to the target prediction step $s$ in the prediction task, while in the classification task, the dimension is set the same as total categories, and a softmax layer is attached at the end of the feedforward networks. The fine-tuning loss settings are also simple. MSE loss is used for the prediction task, and Cross-Entropy loss is used for the classification task.
	
	\section{Experiments}
	\label{Experiment}
	This section conducts various experiments, including downstream task comparison, model hyper-parameter selection, and ablation studies. All the experiments are conducted using an AMAX workstation with Intel(R) Xeon(R) Gold 6226R CPU @ 2.90GHz $\times 64$, RAM 256GB, and RTX3090 $\times 5$.
	
	\subsection{Dataset Settings}
	\label{Dataset Description}
	
	We used a total of 6 real-world open-access datasets to develop a downstream prediction task and an MTS classification task. The first three, namely Appliances Energy Prediction Data(AEPD), Beijing Multi-Site Air-Quality Data(BMAD), and SML2010(SML), are MTS datasets archived in UCI Machine Learning Repository\cite{asuncion2007uci}. These three datasets are used for the prediction-related task. For the classification task, subsets of FaceDetection, JapeneseVowels, and SpokenArabicDigits are chosen from the UEA multivariate time series classification archive\cite{bagnall2018uea}. For training and validation sake, we randomly split each dataset with a ratio of 9:1. Also, we normalize the numerical values for all tasks using the Z-Score method for regular training and fair comparability. We artificially generate missing values for tasks involving missing values under the missing ratio of 5\%, 10\%, and 20\%. 
	
	\begin{table*}[tb]
		\centering
		\small
		\renewcommand\arraystretch{1.2}
        \begin{tabular}{|c|c|ccc|ccc|ccc|c|c|c|}
        \hline
        \multirow{2}{*}{\textbf{Method}}   & \multirow{2}{*}{\textbf{MR}} & \multicolumn{3}{c|}{\textbf{APED}}                                                         & \multicolumn{3}{c|}{\textbf{BMAD}}                                                         & \multicolumn{3}{c|}{\textbf{SML}}                                                          & \multirow{2}{*}{\textbf{UEA-F}} & \multirow{2}{*}{\textbf{UES-S}} & \multirow{2}{*}{\textbf{UEA-J}} \\ \cline{3-11}
                                           &                              & \multicolumn{1}{c|}{\textbf{1s}}    & \multicolumn{1}{c|}{\textbf{3s}}    & \textbf{5s}    & \multicolumn{1}{c|}{\textbf{1s}}    & \multicolumn{1}{c|}{\textbf{3s}}    & \textbf{5s}    & \multicolumn{1}{c|}{\textbf{1s}}    & \multicolumn{1}{c|}{\textbf{3s}}    & \textbf{5s}    &                                 &                                 &                                  \\ \hline
        \multirow{3}{*}{\textbf{LSTNet}}   & \textbf{0.05}                & \multicolumn{1}{c|}{0.136}          & \multicolumn{1}{c|}{0.132}          & 0.137          & \multicolumn{1}{c|}{0.112}          & \multicolumn{1}{c|}{0.106}          & 0.108          & \multicolumn{1}{c|}{0.127}          & \multicolumn{1}{c|}{0.148}          & 0.153          & 72\%                            & 11\%                            & 1\%                              \\ \cline{2-14} 
                                           & \textbf{0.10}                & \multicolumn{1}{c|}{0.158}          & \multicolumn{1}{c|}{0.170}          & 0.180          & \multicolumn{1}{c|}{0.120}          & \multicolumn{1}{c|}{0.135}          & 0.141          & \multicolumn{1}{c|}{0.132}          & \multicolumn{1}{c|}{0.157}          & 0.176          & 70\%                            & 11\%                            & 1\%                              \\ \cline{2-14} 
                                           & \textbf{0.20}                & \multicolumn{1}{c|}{0.197}          & \multicolumn{1}{c|}{0.208}          & 0.216          & \multicolumn{1}{c|}{0.197}          & \multicolumn{1}{c|}{0.221}          & 0.232          & \multicolumn{1}{c|}{0.177}          & \multicolumn{1}{c|}{0.183}          & 0.194          & 65\%                            & 10\%                            & 1\%                              \\ \hline
        \multirow{3}{*}{\textbf{TPA-RNN}}  & \textbf{0.05}                & \multicolumn{1}{c|}{0.185}          & \multicolumn{1}{c|}{0.189}          & 0.219          & \multicolumn{1}{c|}{0.185}          & \multicolumn{1}{c|}{0.215}          & 0.221          & \multicolumn{1}{c|}{0.108}          & \multicolumn{1}{c|}{0.130}          & 0.135          & 69\%                            & 97\%                            & 1\%                              \\ \cline{2-14} 
                                           & \textbf{0.10}                & \multicolumn{1}{c|}{0.197}          & \multicolumn{1}{c|}{0.214}          & 0.225          & \multicolumn{1}{c|}{0.177}          & \multicolumn{1}{c|}{0.225}          & 0.239          & \multicolumn{1}{c|}{0.120}          & \multicolumn{1}{c|}{0.145}          & 0.155          & 69\%                            & 96\%                            & 1\%                              \\ \cline{2-14} 
                                           & \textbf{0.20}                & \multicolumn{1}{c|}{0.215}          & \multicolumn{1}{c|}{0.232}          & 0.248          & \multicolumn{1}{c|}{0.243}          & \multicolumn{1}{c|}{0.264}          & 0.278          & \multicolumn{1}{c|}{0.149}          & \multicolumn{1}{c|}{0.160}          & 0.182          & 63\%                            & 93\%                            & 1\%                              \\ \hline
        \multirow{3}{*}{\textbf{DARNN}}    & \textbf{0.05}                & \multicolumn{1}{c|}{0.129}          & \multicolumn{1}{c|}{0.136}          & 0.129          & \multicolumn{1}{c|}{\textbf{0.067}} & \multicolumn{1}{c|}{0.102}          & 0.105          & \multicolumn{1}{c|}{0.123}          & \multicolumn{1}{c|}{0.127}          & 0.134          & 66\%                            & 64\%                            & 13\%                             \\ \cline{2-14} 
                                           & \textbf{0.10}                & \multicolumn{1}{c|}{0.147}          & \multicolumn{1}{c|}{0.158}          & 0.166          & \multicolumn{1}{c|}{0.087}          & \multicolumn{1}{c|}{0.118}          & 0.124          & \multicolumn{1}{c|}{0.129}          & \multicolumn{1}{c|}{0.137}          & 0.141          & 64\%                            & 63\%                            & 10\%                             \\ \cline{2-14} 
                                           & \textbf{0.20}                & \multicolumn{1}{c|}{0.158}          & \multicolumn{1}{c|}{0.184}          & 0.197          & \multicolumn{1}{c|}{0.094}          & \multicolumn{1}{c|}{0.136}          & 0.144          & \multicolumn{1}{c|}{0.187}          & \multicolumn{1}{c|}{0.161}          & 0.149          & 61\%                            & 56\%                            & 9\%                              \\ \hline
        \multirow{3}{*}{\textbf{DBT-DMAE}} & \textbf{0.05}                & \multicolumn{1}{c|}{\textbf{0.075}} & \multicolumn{1}{c|}{\textbf{0.081}} & \textbf{0.078} & \multicolumn{1}{c|}{0.074}          & \multicolumn{1}{c|}{\textbf{0.094}} & \textbf{0.081} & \multicolumn{1}{c|}{\textbf{0.099}} & \multicolumn{1}{c|}{\textbf{0.100}} & \textbf{0.123} & \textbf{76\%}                   & \textbf{100\%}                  & \textbf{100\%}                   \\ \cline{2-14} 
                                           & \textbf{0.10}                & \multicolumn{1}{c|}{\textbf{0.090}} & \multicolumn{1}{c|}{\textbf{0.071}} & \textbf{0.071} & \multicolumn{1}{c|}{\textbf{0.081}} & \multicolumn{1}{c|}{\textbf{0.088}} & \textbf{0.085} & \multicolumn{1}{c|}{\textbf{0.106}} & \multicolumn{1}{c|}{\textbf{0.123}} & \textbf{0.130} & \textbf{75\%}                   & \textbf{99\%}                   & \textbf{100\%}                   \\ \cline{2-14} 
                                           & \textbf{0.20}                & \multicolumn{1}{c|}{\textbf{0.087}} & \multicolumn{1}{c|}{\textbf{0.088}} & \textbf{0.100} & \multicolumn{1}{c|}{\textbf{0.085}} & \multicolumn{1}{c|}{\textbf{0.097}} & \textbf{0.094} & \multicolumn{1}{c|}{\textbf{0.119}} & \multicolumn{1}{c|}{\textbf{0.109}} & \textbf{0.132} & \textbf{76\%}                   & \textbf{100\%}                  & \textbf{100\%}                   \\ \hline
        \end{tabular}%
		\caption{MAE results on prediction task and precision indicators on classification task}
		\label{result_prediction}
	\end{table*}
	
	\subsection{Performance Indexes}
	There are three types of performance indexes needing to specify.
	\subsubsection{Pre-train Index}
	In this paper, we apply Mean Square Error(MSE) to evaluate the pretrain model's auto-regressive reconstruction performance. We particularly evaluate the both missing and observed values by the performance indexes with $v$ and $m$ suffix. 

	\begin{gather}
	    \label{eq_mse_v}
		\mathrm{MSE}_{v}=\frac{1}{N} \sum_{i=1}^{N} \frac{\Vert M(i)\circ(X(i) - \hat{X}(i))   \Vert_2 }{\Vert M(i) \Vert_2 } \\
        \mathrm{MSE}_{m}=\frac{1}{N} \sum_{i=1}^{N} \frac{\Vert ((\mathbf{1} - M(i)\circ(X(i) - \hat{X}(i)   \Vert_2 }{\Vert \mathbf{1} - M(i) \Vert_2 }
        \label{eq_mse_m}
	\end{gather}
	where $i$ denotes the sample number and $N$ is the total sample number; $\hat{X}$ denotes the reconstructed input $X$; $\Vert\cdot\Vert_{1}$ is the 1-norm operand.
	
	\subsubsection{Prediction Index}
	In terms of prediction-related tasks, we adopt MAE indicator to evaluate the prediction error between the output of downstream fine-tuning and the ground-truth value. The MAE index for $s$ steps prediction is as follows:
	\begin{equation}
	 	\mathrm{MAE}_{p(s)}=\frac{1}{N} \sum_{i=1}^{N} \frac{\Vert X_{T+1:T+s}^{target}(i) - \hat{X}_{T+1:T+s}^{target}(i) \Vert_1 }{s}
	 	\label{eq_predict}
	\end{equation}

	\subsubsection{Classification Index}
	We use the precision indicator for the MTS classification task, which manifests the correctly categorized proportion of samples. The indicator is as follow:
	\begin{equation}
	    \mathrm{PRE}_{c}=\frac{1}{N} \sum_{i=1}^{N} \mathbf{1}(c(i)=\hat{c}(i))  
	    \label{eq_precision}
	\end{equation}
	where $\mathbf{1}(\cdot)$ is the indicator function that returns one if the condition is true; otherwise, it returns 0.
	\subsection{Baselines Settings}
	
	In terms of prediction baselines, we choose three different State of the Art models, namely LSTNet\cite{LSTNet}, TPA-RNN\cite{TPA-RNN} and DARNN\cite{DARNN}. In recent literature, these three models are well-recognized sequential prediction and estimation methods using deep learning. For the classification task, we append a softmax layer at the rear of the three models above; therefore, the output is the probability under which MTS input belongs to each category.

	\subsection{Comparative Study}
	\label{sec:comparative study}
	This subsection compares DBT-DMAE's prediction and fine-tuning classification performances with the three baselines above. The missing ratio(MR) is artificially set to 5\%, 10\%, and 20\% for each dataset. The results in the first three datasets manifest the prediction performances computed as Equal.(\ref{eq_predict}), and the 1s, 3s, and 5s denote 1-step, 3-step, and 5-step prediction, respectively. The last three columns show the comparative results of the classification task, the precision indexes are computed as Equal.(\ref{eq_precision}). The best performance indicator in each case is written in bold. According to the results, fine-tuning of DBT-DMAE outperforms the three compared baselines, especially when the missing ratio is high.

    \subsection{Hyper-Parameter Selection}
	\label{hyper}
	In this subsection, we conduct experiments to discuss the DBT-DMAE's hyper-parameters effect on model convergence and performances. Three crucial parameters are taken into account: hidden states number $h_s$, random masking ratio $m_r$, and kernel size combination $ks$. We develop a grid search of the hyper-parameter fields. The $h_s$ are selected from [16, 32, 64, 128]; $m_r$ is chosen from [0.05, 0.1, 0.2, 0.3]; $ks$ is chosen from [\{3, 5, 7\}, \{2, 3, 5\}, \{5, 7, 11\}, \{2, 3, 7\}]. For brevity, we only present the performance using the SML dataset here.
	\begin{figure*}[ht]
		\centering
		\subfigtopskip=-5pt
		% \subfigbottomskip=2pt
		% \subfigcapskip=-5pt
		\subfigure[loss curve with different $h_s$ s]{
			{\includegraphics[width=0.3\textwidth]{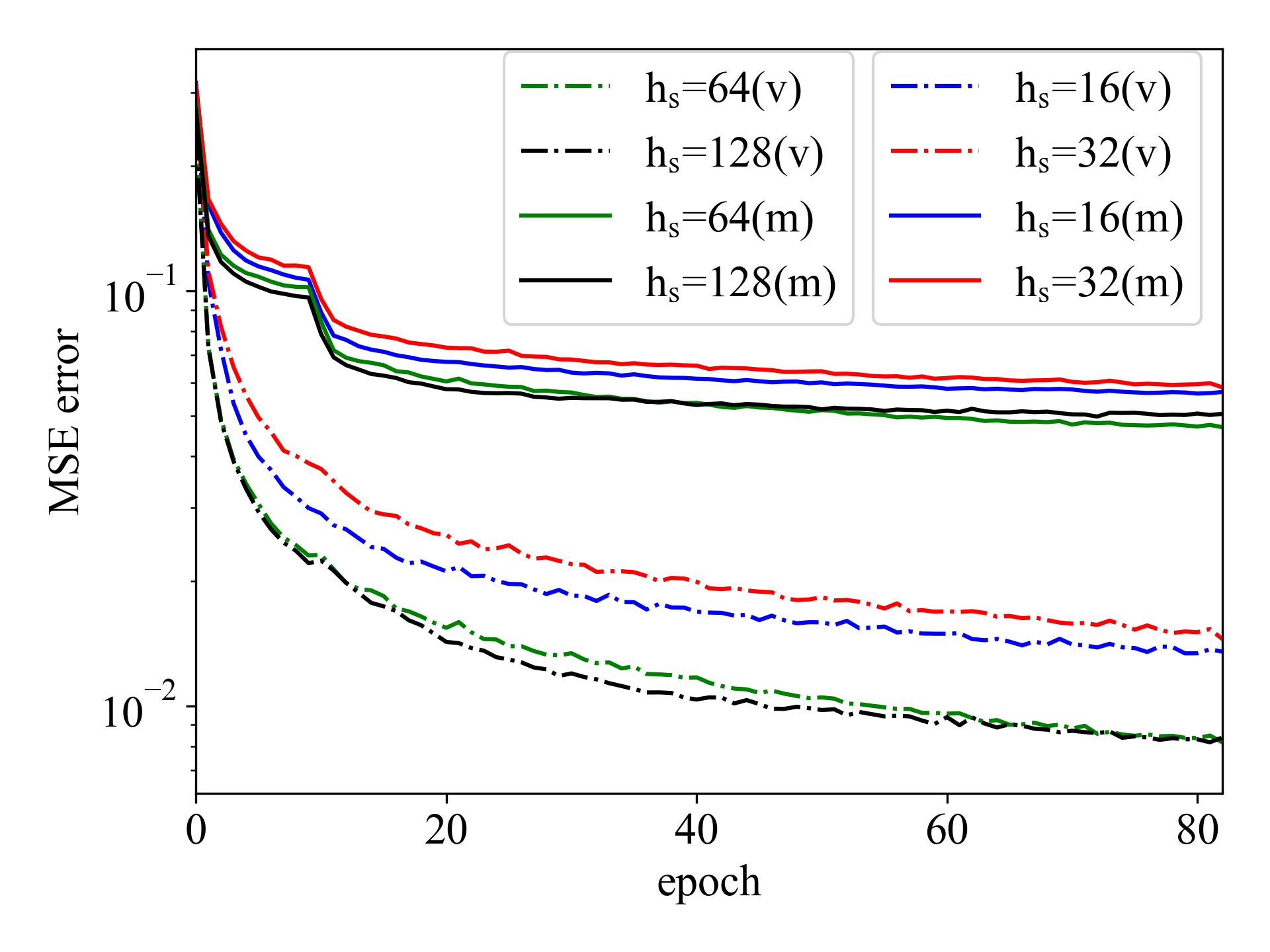}}}
		\subfigure[loss curve with different $m_r$ s]{
			{\includegraphics[width=0.3\textwidth]{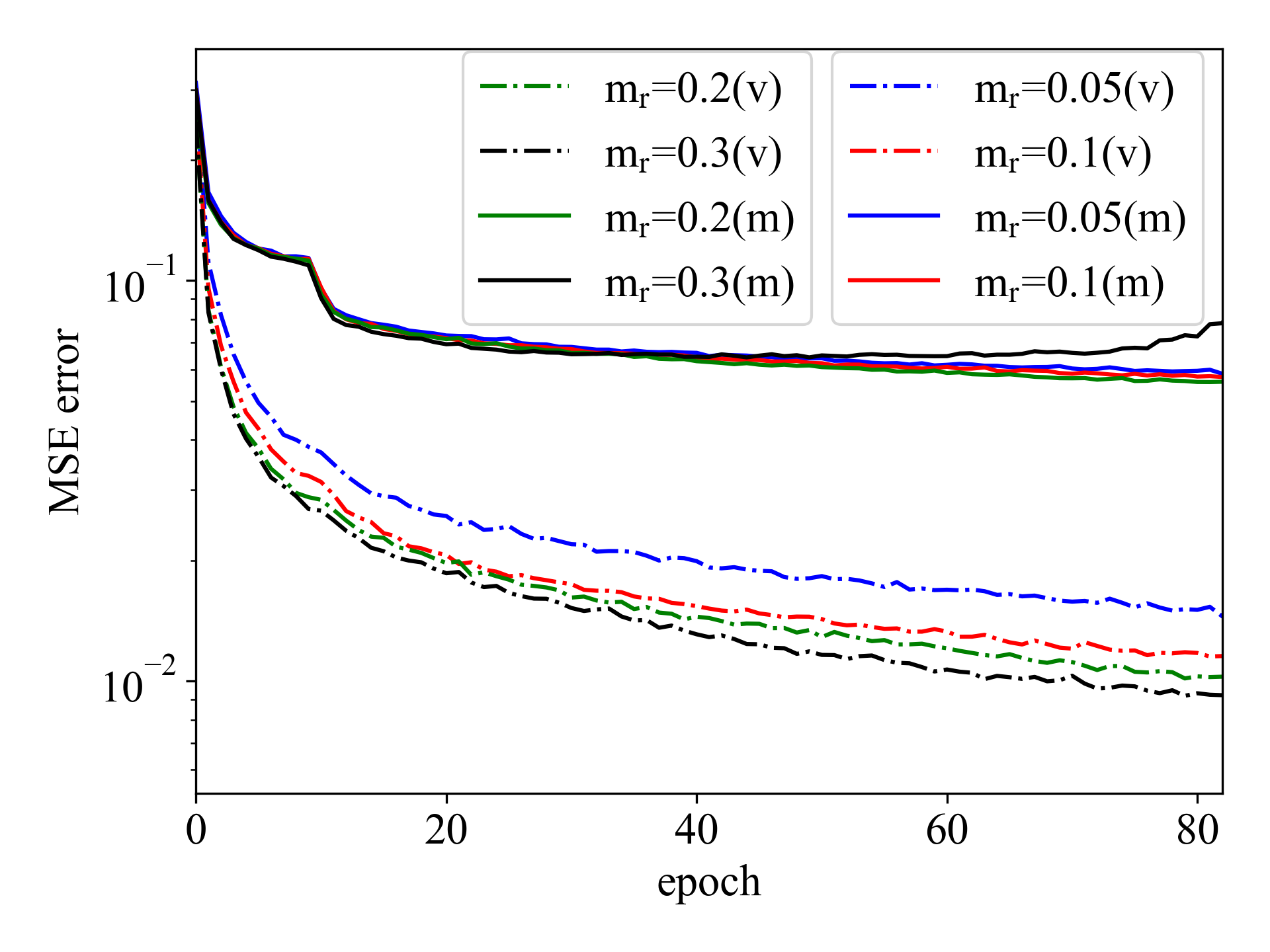}}}
		\subfigure[loss curve with different $ks$ s]{
			{\includegraphics[width=0.3\textwidth]{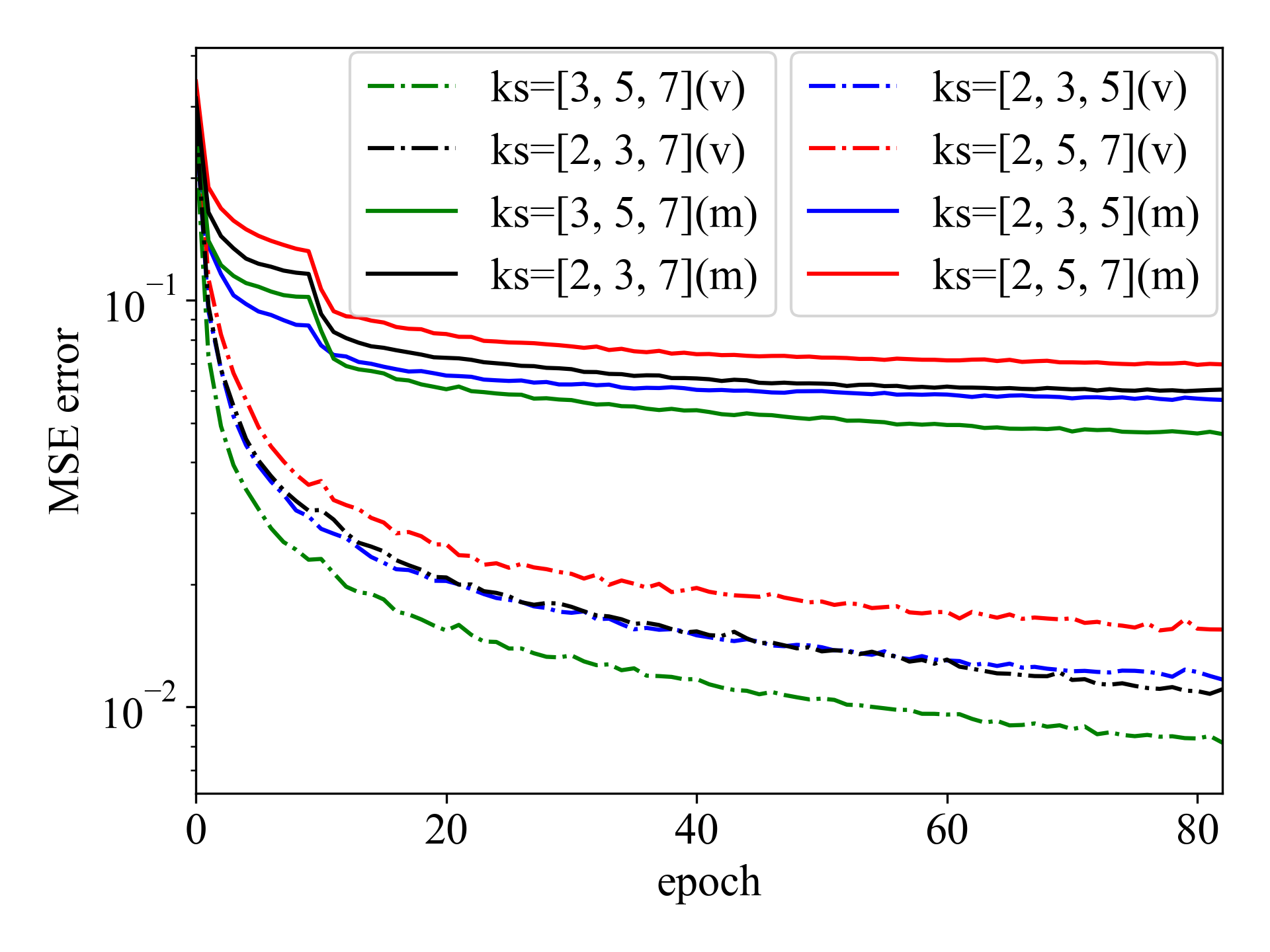}}}
		
		\caption{MSE loss curves at both observed and missing entries with different $h_s$, $m_r$, and $ks$ combinations}
		\label{f_hyper}
	\end{figure*}
	
	In Figure.\ref{f_hyper}, we visualize the grid search results containing training and validation MSE indexes under different hyper-parameter combinations. Specifically, the restoration performances at missing and observed entries are plotted separately, and the labels are marked with (v) and (m), respectively. It is worth mentioning that there is a rapid loss descent after epoch 10, which accurately reflects the effect of the warm-up training trick. In conclusion, in order to balance the model's performance, converge speed and model volume, we finally choose the hyper-parameter combination that $m_r=0.2, h_s=64, ks=\{3, 5, 7\}$.
	
	\subsection{Ablation Studies}
	\label{subsec:ablation studies}
	Our algorithm has four critical sub-structures: DPE, RM, DBT, and ASF. The ablation studies are delivered in four independent parts further to verify the efficiency and brevity of our model:
	\begin{enumerate}
		\item w/o DPE: substitute DPE by the hard-code embedding.
		\item w/o RM:  remove random masking process and first part of the loss function.
		\item w/o DK:  substituted DBT unit by the original TCN.
		\item w/o ASF: substituted ASF by linear concatenation.
	\end{enumerate}
	All the ablation experiment results below are obtained under the AEPD dataset and missing ratio equal to 0.2.

	\begin{table}[htb]

		\small
		\centering
		\renewcommand{\arraystretch}{1.2}
        \begin{tabular}{|c|cccc|}
        \hline
        \multirow{2}{*}{\textbf{Case}} & \multicolumn{3}{c|}{\textbf{Evaluation Metrics}}                               \\ \cline{2-4} 
                                       & \multicolumn{1}{c|}{$\mathbf{MSE_{v}}$} & \multicolumn{1}{c|}{$\mathbf{MSE_{m}}$} & \multicolumn{1}{c|}{$\mathbf{MAE_{p(3)}}$}  \\ \hline
        \textbf{DBT-DMAE}              & \multicolumn{1}{c|}{\textbf{0.016}} & \multicolumn{1}{c|}{\textbf{0.021}} & \multicolumn{1}{c|}{\textbf{0.088}}          \\ \hline
        \textbf{w/o DPE}               & \multicolumn{1}{c|}{0.023}          & \multicolumn{1}{c|}{0.032}          & \multicolumn{1}{c|}{0.096}          \\ \hline
        \textbf{w/o RM}               & \multicolumn{1}{c|}{0.028}          & \multicolumn{1}{c|}{0.160}          & \multicolumn{1}{c|}{0.124}           \\ \hline
        \textbf{w/o DK}                & \multicolumn{1}{c|}{0.019}          & \multicolumn{1}{c|}{0.025}          & \multicolumn{1}{c|}{0.101}          \\ \hline
        \textbf{w/o ASF}               & \multicolumn{1}{c|}{0.021}          & \multicolumn{1}{c|}{0.024}          & \multicolumn{1}{c|}{0.092}          \\ \hline
        \end{tabular}%
		\caption{Ablation Studies Result}
		\label{t_ablation}
	\end{table}
	
	Table.\ref{t_ablation} shows the ablation experiments' result. We evaluate the pre-train performance within both observed and missing locations and the downstream 3-step prediction precision index is also included for support. 
	
	The results from cases w/o DPE and w/o RM manifest that the dynamic missing embedding is exceedingly meaningful in the whole algorithm. When the random masking process is removed from the missing representation procedure, the reconstruction performance on the missing part dramatically drops. In contrast, the performances in the observed parts remain unaffected. This unbalanced performance degradation can be viewed as a failure signal in the missing value representation. 
	
\begin{figure}[htb]
		\centering
		\subfigtopskip=-5pt
		% \subfigbottomskip=2pt
		% \subfigcapskip=-5pt
		\subfigure[DPE along with the time axis]{\includegraphics[width=0.8\columnwidth]{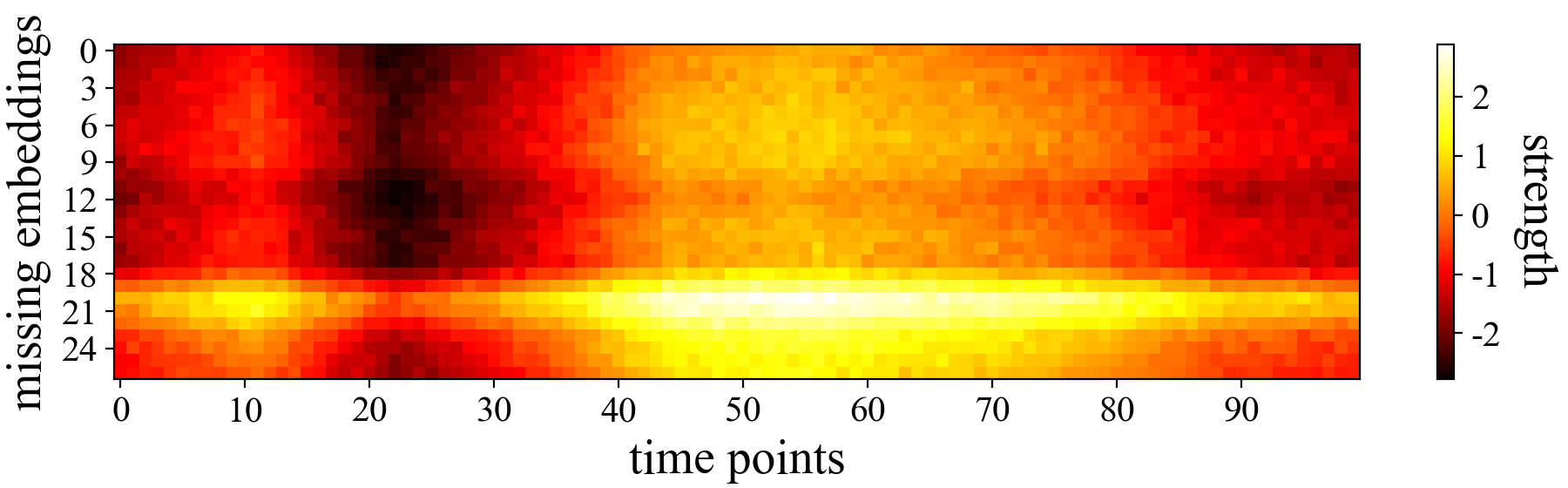}}
		\subfigure[DK parameters along with the time axis]{\includegraphics[width=0.8\columnwidth]{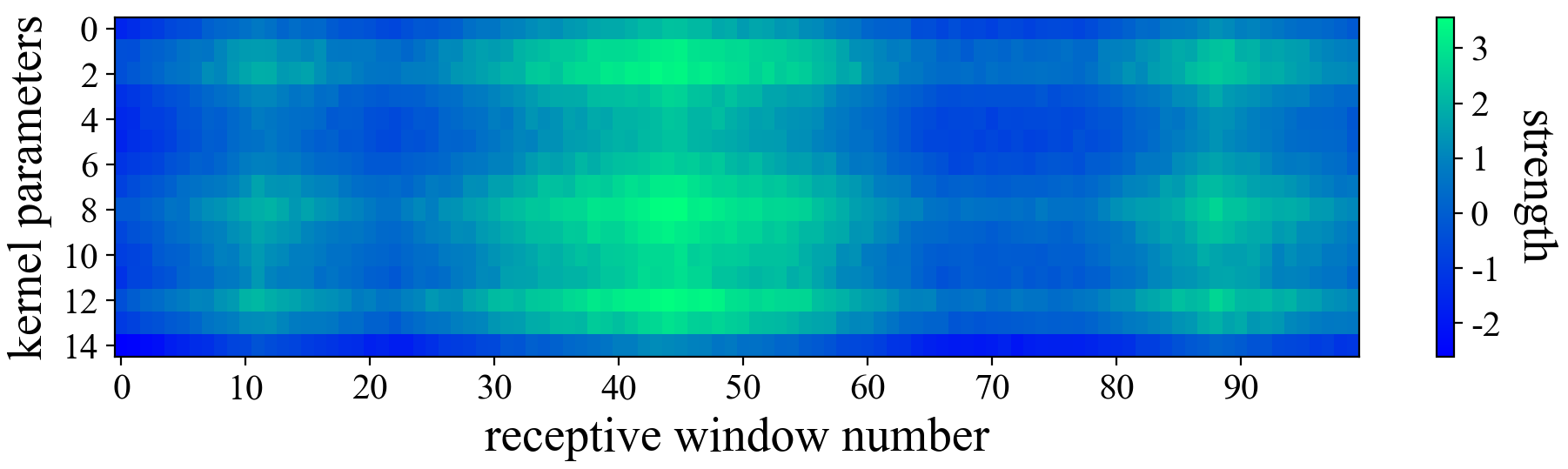}}
		
		\caption{Visualization of DBT-DMAE's dynamic characteristics.}
		\label{f_D}
	\end{figure}

	In terms of experiment cases w/o DK and w/o ASF, ablation results confirm these substructures' dynamic feature extraction ability. We also visualize the dynamic kernel parameters and dynamic missing positional embedding values in Figure \ref{f_D}, which also confirms the validity of the proposed DBT structure.
	
	\section{Conclusion}
	\label{Conclusion}
	In this paper, we propose a universal MTS pre-train model to obtain  downstream-generalized encoded representation conquering the missing data problem. Our  model utilizes the proposed DBT as the basic unit, and adopts the dynamic positional embedding and mask-learning mechanism to construct the auto-regressive DMAE. With simple fine-tuning, the proposed model is suitable for various downstream tasks, including prediction, classification, condition estimation, etc. Various experiments with open real-world datasets manifest the superiority of the proposed DBT-DMAE when confronting the missing data problem in the MTS context.  
    \newpage

% Use \bibliography{yourbibfile} instead or the References section will not appear in your paper
\bibliography{references}

\end{document}